\documentclass[runningheads]{llncs}

\usepackage{eccv}

\usepackage{eccvabbrv}

\usepackage{graphicx}
\usepackage{booktabs}

\usepackage[accsupp]{axessibility}  

\usepackage{amsmath,amssymb,amsfonts}
\usepackage{commath}
\usepackage{derivative}
\usepackage{subcaption}
\usepackage[export]{adjustbox}
\usepackage{multirow}
\usepackage{booktabs}
\DeclareMathOperator*{\argmin}{\text{arg}\min}

\usepackage{hyperref}

\usepackage{orcidlink}

\begin{document}

\title{LiFlow: Flow Matching for 3D LiDAR Scene Completion} 


\author{Andrea Matteazzi \and
Dietmar Tutsch}

\authorrunning{A.~Matteazzi and D.~Tutsch}

\institute{University of Wuppertal, Wuppertal, Germany \\
\email{\{matteazzi,tutsch\}@uni-wuppertal.de}}

\maketitle

\begin{abstract}
\begin{figure*}[t]
\centering
\includegraphics[width=1.0\linewidth]{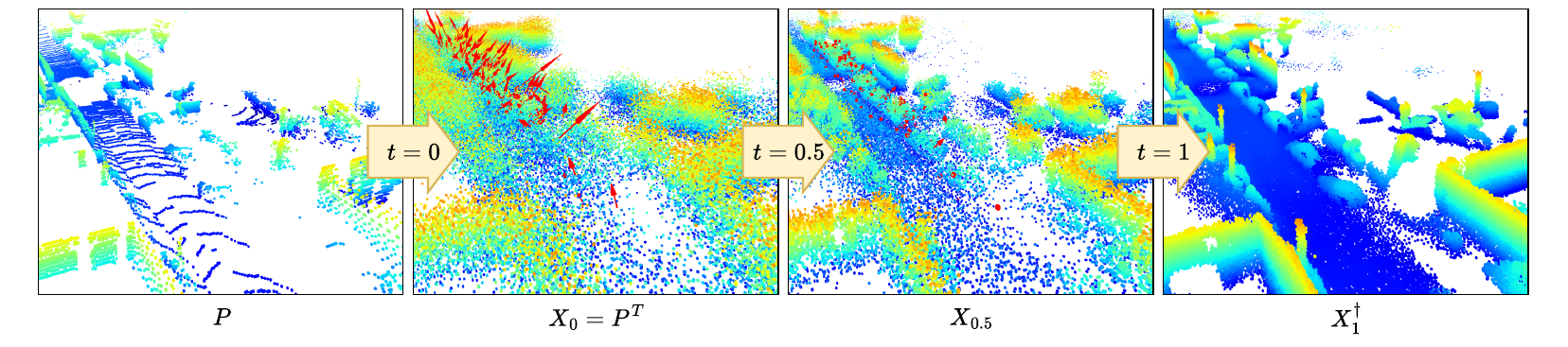}
\caption{Starting from a single LiDAR Scan $P$ from the validation set of the SemanticKITTI dataset (sequence $08$), we compute the initial point cloud $P^T$. LiFlow generates a flow of point clouds with Euler’s method~\cite{lipman2023flow} by predicting time-dependent vector fields $u_\theta^*(t,X_t, P)$ in continuous time $t: 0 \rightarrow 1$, achieving a 3D complete scene. The arrows indicate scaled version of $u_\theta^*(t,X_t, P)$ for some points labeled in $P$ as car. Colors depict point height. $\dagger$: with refinement network~\cite{nunes2024scaling}.}
\label{fig:01}
\end{figure*}
In autonomous driving scenarios, the collected LiDAR point clouds can be challenged by occlusion and long-range sparsity, limiting the perception of autonomous driving systems.
Scene completion methods can infer the missing parts of incomplete 3D LiDAR scenes.
Recent methods adopt local point-level denoising diffusion probabilistic models, which require predicting Gaussian noise, leading to a mismatch between training and inference initial distributions. This paper introduces the first flow matching framework for 3D LiDAR scene completion, improving upon diffusion-based methods by ensuring consistent initial distributions between training and inference. 
The model employs a nearest neighbor flow matching loss and a Chamfer distance loss to enhance both local structure and global coverage in the alignment of point clouds. 
LiFlow achieves state-of-the-art performance across multiple metrics. Code:~\url{https://github.com/matteandre/LiFlow}.
\end{abstract}

\section{Introduction}
\label{sec:intro}

LiDAR scenes constitute a fundamental source of data for perception modules, enabling autonomous driving systems to perceive and understand the surrounding environment. LiDAR sensors provide 3D information of the vehicle surroundings in the form of 3D point clouds.
The reliability of LiDAR-based perception modules is therefore intrinsic to the accuracy and completeness of the 3D point clouds. Despite the accuracy of LiDAR sensors, in autonomous driving scenarios, the collected point clouds are affected by occlusion and long-range sparsity. The resulting gaps in the collected point clouds can hinder perception systems from perceiving objects and structures in the environment. Scene completion methods can infer the missing parts of incomplete 3D scenes. Providing a dense and more complete scene representation of non-observed regions can add valuable information to autonomous driving systems, helping to improve different tasks such as object detection~\cite{xiong2023learning}, semantic segmentation~\cite{matteazzi2025preprocessing},
localization, and navigation.
Recent works~\cite{nunes2024scaling,martyniuk2025lidpm} use local point-level Denoising Diffusion Probabilistic Models (DDPMs)~\cite{ho2020denoising} for 3D LiDAR scene completion at scene scale. While effective, DDPMs are limited by their need to infer Gaussian noise, creating a mismatch between the initial point cloud distributions at training and inference. We introduce LiFlow, the first Flow Matching (FM)~\cite{lipman2023flow} formulation for 3D LiDAR scene completion (\cref{fig:01}). By enabling flexible distribution matching without assuming Gaussian relations, LiFlow ensures consistent initial distributions between training and inference.

We compare LiFlow with different scene completion methods and conduct extensive experiments to validate our proposed 3D flow matching formulation.
In summary, our key contributions are:
\begin{itemize}
\item We formulate the first flow matching paradigm for 3D LiDAR scene completion.
\item We solve the mismatch of initial point cloud distribution with novel point-wise nearest neighbor flow matching and Chamfer distance losses.
\item Our model achieves state-of-the-art results compared to previous diffusion and non-diffusion methods.
\end{itemize}

\section{Related Work}
\subsection{Scene Completion}
Scene completion is the task of inferring the missing parts of an incomplete 3D scene provided by a sensor. Given the LiDAR data sparsity, providing a dense and more complete scene representation of non-observed regions is fundamental to enlarge the incomplete data measured by the sensor and add valuable information to perception modules.
Early works~\cite{jaritz2018sparse,ma2018sparse,ma2019self,fu2020depth} propose scene completion as a prediction of dense depth images extracted from RGB images and fused with LiDAR point clouds.
Recent approaches address the scene completion task in the 3D space for denser completions.
Some methods~\cite{vizzo2022make,li2023lode} leverage Signed Distance Fields (SDF) to represent 3D LiDAR scenes as voxel grids where each voxel stores the distance to the nearest surface, and they train a 3D convolutional network to predict such SDF volumes from LiDAR scenes. Besides, further methods operating at the voxel level~\cite{roldao2020lightweight,yan2021sparse,xia2023scpnet} aim to solve semantic scene completion, leveraging and inferring also semantic labels for each voxel.
The scene completion accuracy of methods operating at the voxel level is intrinsic to the voxel grid resolution.
Denoising Diffusion Probabilistic Models (DDPM)~\cite{ho2020denoising} show promising results in the generation of 3D point clouds of single object shapes~\cite{luo2021diffusion,Zhou_2021_ICCV,lee2023diffusion} and more recently in the completion of 3D LiDAR scenes~\cite{nunes2024scaling,martyniuk2025lidpm} at the point level.

\subsection{Diffusion Models for 3D Scene Completion}
\label{subsec:diff}

Recent works~\cite{nunes2024scaling,martyniuk2025lidpm} formulate a diffusion-based~\cite{ho2020denoising} procedure for 3D LiDAR scene completion at the point level.
Given a complete 3D LiDAR scene $G \in \mathbb{R}^{M \times 3}$ and a single LiDAR scan $P \in \mathbb{R}^{N\times 3}$, LiDiff~\cite{nunes2024scaling} defines a forward diffusion process by adding local noise independently to each point $\mathbf{p}_m \in G$:
\begin{equation}
    \mathbf{p}^t_m = \mathbf{p}_m + \sqrt{1-\bar{\alpha_t}}\varepsilon, \quad  \varepsilon \sim \mathcal{N}(0,I),\\
    \label{eq:1}
\end{equation}
with $\bar{\alpha_t}=\prod_{i = 1}^{t} \alpha_i$, $\alpha_t=1-\beta_t$, and $\beta_1, ..., \beta_T$ noise factors parameterizing the diffusion process.
At each diffusion step $t$, the model is trained to predict the noise $\varepsilon$ and progressively denoise the point set $G^t$ toward the target scene $G$.
\Cref{eq:1} implicitly enforces a priori Gaussian point-wise correspondence between $G$ and $G^t$, as each noisy point $\mathbf{p}^t_m$ remains associated with the same original point $\mathbf{p}_m$.
At each iteration, LiDiff is trained by sampling a random step $t \in T$ and computing $G^t$ from $G$ with \cref{eq:1}.

\noindent
During inference, due to this local noise diffusion formulation, $G^T$ is not available.
Instead, $G^T$ is approximated from $P$. By concatenating its points $K$ times such that $M=KN$, they generate $P^* \in \mathbb{R}^{M \times 3}$ and compute the initial noisy point cloud $P^T$ from $P^*$ with \cref{eq:1}.
This mismatch of initial point cloud distribution between training and inference can hinder the denoising process as the sparsity of single LiDAR scans $P$, especially in the presence of occlusion and in long distances, can limit the coverage of particular structures in $G$.

\subsection{Flow Matching for 3D Data}
While Denoising Diffusion Probabilistic Model (DDPM) represents a discrete time formulation of diffusion and denoising, Song \etal~\cite{song2021scorebased} prove its equivalent formulation in continuous time. In continuous time, denoising steps are equivalent to solving a Stochastic Differential Equation (SDE)~\cite{NEURIPS2022_260a14ac}. The more general Flow Matching (FM)~\cite{lipman2023flow} paradigm extends the continuous Diffusion formulation with an equivalent Ordinary Differential Equation (ODE) formulation not restricted to Gaussian distributions.
FM demonstrates great improvements over DDPM in different image generation datasets~\cite{lipman2023flow,schusterbauer2025diff2flow}. FM formulation allows mapping input and target distributions not restricted to Gaussian distributions and permits faster convergence by not introducing stochasticity during the matching procedure~\cite{schusterbauer2024fmboost}.
Recent works~\cite{liu2025efficient,lan2025gaussiananythinginteractivepointcloud} leverage FM for the generation of 3D point clouds of single object shapes. Liu, He, and Li ~\cite{liu2025efficient} propose to upsample 3D point clouds by introducing a pre-alignment method based on Earth Mover’s Distance (EMD) optimization to ensure coherent interpolation between sparse and dense point clouds, and they formulate FM with a straight path~\cite{lipman2023flow}.

\section{Preliminary}
\subsection{Flow Matching}
Flow Matching (FM)~\cite{lipman2023flow} is a paradigm for generative modeling that regresses vector fields to generate target probability density paths. 
Let $x \in \mathbb{R}^d$ be the data points. We can define a probability density path $p_t:  [0,1]\times \mathbb{R}^d \rightarrow \mathbb{R}_{>0}$, as a time-dependent probability density function, \ie, $\int p_t(x) dx = 1$ and a time-dependent vector field $v_t: [0,1] \times \mathbb{R}^d \rightarrow \mathbb{R}^d$. A vector field $v_t$ can be used to construct a flow  $\phi_t: [0, 1] \times \mathbb{R}^d \rightarrow \mathbb{R}^d$, defined as the solution of the ordinary differential equation (ODE): 
\begin{equation}
\begin{aligned}
& \phi_0(x) = x \text{,} \\
& \odv{\phi_t(x)}{t} = v_t(\phi_t(x)) \text{.}
\end{aligned}
 \label{eq:2}
\end{equation}

\noindent
A vector field $v_t$ generates a probability density path $p_t$ if its flow $\phi_t$ satisfies the push-forward equation:
\begin{equation}
p_t = [\phi_t(x)]p_0 = p_0(\phi_t^{-1}(x))det\left[\frac{\partial \phi_t^{-1}(x)}{\partial x}\right] \text{.}
 \label{eq:3}
\end{equation}

\noindent
Let $x_1 \sim q(x_1)$ be data points distributed according to the data distribution $q$. We assume to only have access to data samples from $q(x_1)$ but have no access to the density function itself. We let $p_t$ be a probability density path such that $p_0 = p$ is a simple initial distribution, and let $p_1$ be approximately equal in distribution to $q$. The FM objective is designed to match this target probability path, which will allow us to flow from $p_0$ to $p_1$. Given a target vector field $v_t(x)$ generating this target probability density path $p_t$ \ie, $x_t = \phi_t(x_0) \sim p_t$, $x_0 \sim p_0$, we can regress a vector field $u_\theta(t,x)$ parameterized by a neural network with learnable parameters $\theta$ using the FM objective as:

\begin{equation}
    \mathcal{L}_{FM}(t,x_0) = \norm{u_\theta(t,\phi_t(x_0)) - v_t(\phi_t(x_0))}^2 \text{.}
     \label{eq:4}
\end{equation}

\noindent
Since we generally do not have access to a closed form of $v_t$, we can acquire the same gradients and therefore efficiently regress the neural network using the Conditional Flow Matching (CFM) objective:

\begin{equation}
    \mathcal{L}_{CFM}(t,x_0,x_1) =\norm{u_\theta(t,\phi_t(x_0|x_1)) - v_t(x_0|x_1)}^2 \text{,}
     \label{eq:5}
\end{equation}
\noindent
with $\phi_t(x_0|x_1)$ conditional flow and $v_t(x_0|x_1)$ conditional vector field.

\subsection{Data-Dependent Couplings}
The data-dependent couplings~\cite{schusterbauer2024fmboost,tong2024improving} formulation allows us to create a probability path between an initial signal and target data within the Flow Matching (FM) objective.
Let $x_1$ be target data and $x_0$ be an initial signal obtained from $x_1$. 
We can smooth around the data samples within a minimal variance to acquire the corresponding probability distribution $p_0=\mathcal{N}(x_0, \sigma^2_{min}I)$, $p_1=\mathcal{N}(x_1, \sigma^2_{min}I)$. The flow can be associated with Optimal Transport (OT) and defined by the equations:

\begin{equation}
    \begin{aligned}
         &\phi^C_t(x_0|x_1) = tx_1 + (1-t)x_0 \text{,} \\
         &v^C_t(x_0|x_1) = x_1 - x_0  \text{.}
    \end{aligned}
     \label{eq:6}
\end{equation}

The corresponding conditional probability path is:
\begin{equation}
    p_t(x_t|x_1) = \mathcal{N}(x_t|tx_1 + (1-t)x_0, \sigma^2_{min}I) \text{.}
     \label{eq:7}
\end{equation}

The resulting Data-Dependent Coupling Flow Matching (DFM) loss becomes:

\begin{equation}
    \mathcal{L}_{DFM}(t,x_0,x_1) = \norm{u_\theta(t, \phi^C_t(x_0|x_1)) - (x_1 - x_0)}^2\text{.}
     \label{eq:8}
\end{equation}

\noindent

\section{Method}
We present LiFlow as the first Flow Matching (FM)~\cite{lipman2023flow} paradigm that aims to solve the mismatch of initial point cloud distribution between training and inference of LiDiff~\cite{nunes2024scaling} analyzed previously.
Thanks to the flexibility of FM, we can use the same initial point cloud distribution for both the training and inference.
\subsection{Flow Matching for 3D Scene Completion}
Let $G \in \mathbb{R}^{M \times 3}$ be a 3D LiDAR complete scene and $P \in \mathbb{R}^{N\times 3}$ a single LiDAR scan. Similarly to LiDiff~\cite{nunes2024scaling}, we generate $P^* \in \mathbb{R}^{M \times 3}$ from $P$ by concatenating its points $K$ times such that $M=KN$. Furthermore, similar to \cref{eq:1}, we compute the initial noisy point cloud $P^T$ from $P^*$ as a noise offset added locally to each point $\mathbf{p}^* \in P^*$:
\begin{equation}
    \mathbf{p}^T = \mathbf{p}^* +\varepsilon\text{,} \quad \varepsilon \sim \mathcal{N}(0,I) \text{.}
     \label{eq:9}
\end{equation}

\noindent
Following the data-dependent couplings~\cite{schusterbauer2024fmboost,tong2024improving}, we set the target data $x_1=G$ and the initial signal $x_0=P^T$.

This formulation enables training and inference from the same initial distribution, addressing the mismatch issue of LiDiff. However, a point-wise correspondence between $x_0$ and $x_1$ is not yet established, as the points originate from different clouds.

The training is implemented with classifier-free guidance~\cite{ho2021classifierfree} with condition $c$ associated with the single LiDAR scan $P$ \ie, $\tilde{c}\sim \mathbb{B}(p)$ as a Bernoulli distribution of outcomes $\{\emptyset, P\}$ with probability $p$ that $\emptyset$ occurs.

\subsection{Nearest Neighbor Flow Matching}

To satisfy \cref{eq:6}, we introduce point-wise Nearest Neighbor Flow Matching (NFM) to align a noisy point cloud $x_0$ with the target $x_1$. NFM establishes point-wise correspondences by assigning each point in $x_0$ to its nearest neighbor in $x_1$. Unlike LiDiff~\cite{nunes2024scaling}, which assumes Gaussian point-wise correspondences a priori, NFM adapts a posteriori to the initial distribution $P^T$, enabling consistent initial point cloud distributions during training and inference. Formally, the Nearest Neighbor (NN) of each point in $x_0$ is defined as the point in $x_1$ minimizing the $\mathcal{L}_2$ distance over 3D coordinates:
\begin{equation}
    NN(P,\hat{P}) = \bigg(\argmin_{\hat{\mathbf{p}} \in \hat{P}} \norm{\mathbf{p} - \hat{\mathbf{p}}} \bigg)_{\mathbf{p} \in P}\text{.}
    \label{eq:10}
\end{equation}
With our definition, the flow formulation in \cref{eq:6} becomes:
\begin{equation}
    \begin{aligned}
         &\phi^{N}_t(x_0|x_1) = tNN(x_0,x_1) + (1-t)x_0 \text{,} \\
         &v^{N}_t(x_0|x_1) = NN(x_0,x_1) - x_0 \text{,}
    \end{aligned}
     \label{eq:11}
\end{equation}

and the loss in \cref{eq:8} becomes:

\begin{equation}
     \mathcal{L}_{NFM}(t,x_0,x_1,\tilde{c}) = \norm{u_\theta(t, \phi^{N}_t(x_0|x_1), \tilde{c}) - v^{N}_t(x_0|x_1)}^2 \text{,}
     \label{eq:12}
\end{equation}

\noindent
with $u_\theta$ MinkUNet~\cite{choy20194d} model.

\subsection{Chamfer Distance Matching}

Nearest Neighbor Flow Matching (NFM) does not ensure one-to-one correspondences, as multiple points in $x_0$ may map to the same point in $x_1$. This can produce discrete reconstructions with reduced scene occupancy and lower fine-grained point density. To address this, we introduce the Chamfer Distance Matching (CDM) loss based on the target objective in \cref{eq:8}:

\begin{equation}
    \mathcal{L}_{CDM}(t,x_0,x_1,\tilde{c}) =  CD(x_0 + u_\theta(t, \phi^{N}_t(x_0|x_1), \tilde{c}), x_1) \text{,}
    \label{eq:13}
\end{equation}

\noindent
where the Chamfer Distance (CD) is defined as:
\begin{equation}
    CD(P,\hat{P})=\sum_{\mathbf{p} \in P} \min_{\hat{\mathbf{p}} \in \hat{P}} \norm{\mathbf{p} - \hat{\mathbf{p}}}^2 + \sum_{\hat{\mathbf{p}} \in \hat{P}} \min_{\mathbf{p} \in P} \norm{\mathbf{p} - \hat{\mathbf{p}}}^2\text{,}
    \label{eq:14}
\end{equation}

\noindent
and assures a point-wise matching between $x_0$ and $x_1$.
CDM encourages the model to predict vector fields that spread out the generated point cloud to cover the entire target scene by accounting for mutual alignment between $x_0$ and $x_1$.

\subsection{Training and Inference}

\subsubsection{Training}
During training, given a target data $x_1$ and a condition $c$, we compute $x_0$ with \cref{eq:9} and uniformly sample $t \in [0,1]$. Following \cref{eq:11}, we compute the conditional flow and the vector field, and compute the final loss:
\begin{equation}
\mathcal{L}(t, \mathbf{x}_0, \mathbf{x}_1, \tilde{c}) = \lambda_{\text{NFM}}\mathcal{L}_{NFM} + \lambda_{\text{CDM}}\mathcal{L}_{CDM},
\label{eq:15}
\end{equation}

\noindent
where $\mathcal{L}_{NFM}$ establishes the point-wise flow matching, and $\mathcal{L}_{CDM}$ encourages the generated point cloud to spread out and align with the target complete scene. 
$\lambda_{\text{NFM}}$ and $\lambda_{\text{CDM}}$ represent the respective loss weights.

\subsubsection{Inference}

During inference, given a single LiDAR scan $P$, we compute $x_0$ with \cref{eq:9}, and we progressively follow the flow $t: 0 \rightarrow 1$ using the vector field predicted by the model using Euler's method~\cite{lipman2023flow}:
\begin{equation}
\begin{aligned}
    &X_0 = x_0\text{,} \\
    &X_{t+h} = X_t + h u_\theta^*(t,X_t, P) \; (t= 0, h, 2h, ..., 1-h) \text{,}
    \end{aligned}
    \label{eq:16}
\end{equation}

\noindent
where $h$ is the step size and $u_\theta^*(t,X_t, P)$ is obtained through classifier-free guidance~\cite{ho2021classifierfree}:
\begin{equation}
    u_\theta^*(t,X_t, P) = u_\theta(t,X_t, \emptyset) + w[u_\theta(t,X_t, P) - u_\theta(t,X_t, \emptyset)] \text{,}
    \label{eq:17}
\end{equation}

\noindent
with $w$ the classifier-free conditioning weight~\cite{ho2021classifierfree}.

\section{Experiments}

\subsection{Experimental Settings}
\subsubsection{Datasets}
For the following experiments, we train LiFlow on the SemanticKITTI~\cite{behley2019semantickitti} dataset of training sequences $00-07$ and $09-10$ and use sequence $08$ for validation.
Besides, we also evaluate and compare our model on the Apollo Columbia Park MapData~\cite{lu2019l3} dataset sequence $00$. This dataset is recorded with the same sensor, but in a different environment.
For all baselines, we use the provided weights also trained on SemanticKITTI and their default parameters.
We follow Nunes \etal~\cite{nunes2024scaling} to generate the ground truth complete scans, using the dataset poses to aggregate the scans in the sequence and remove moving objects to build a map for each sequence. Moving objects are removed using the ground truth semantic labels for SemanticKITTI and the labels provided by Chen \etal; Mersch
\etal~\cite{chen2022automatic,mersch2023building} for the Apollo dataset.

\subsubsection{Training}
We train LiFlow for $20$ epochs, using only the training set from SemanticKITTI.
We set the classifier-free probability~\cite{ho2021classifierfree} $p = 0.1$.
The loss in \cref{eq:15} is implemented with $\lambda_{\text{NFM}}=1$ and $\lambda_{\text{CDM}}=0.1$.
For the MinkUNet~\cite{choy20194d} $u_\theta$, we follow Martyniuk \etal ~\cite{martyniuk2025lidpm} and we replace Batch Normalization (BN)~\cite{ioffe2015batch} layers with Instance Normalization (IN)~\cite{ulyanov2016instance} layers from the architecture proposed by Nunes \etal~\cite{nunes2024scaling}.
Besides, we set the quantization resolution to $0.05$~m and we train $u_\theta$ with Exponential Moving Average (EMA)~\cite{tarvainen2017mean} with decay $\alpha=0.9999$.
The training optimization is implemented with Adam~\cite{adam} and the batch size is set to $4$.
For each input scan, we define the scan range as $50$~m and sample $N = 18, 000$ points with farthest point sampling. For the ground truth, we randomly sample $M = 180, 000$ points without replacement, \ie, $K=10$.
Training and inference are performed using an NVIDIA A100 GPU with $80$ GB of memory.

\subsubsection{Inference}
For inference, we use Euler's method for $10$ steps, \ie, $h=0.1$. We set the classifier-free conditioning weight~\cite{ho2021classifierfree} to $w = 6.0$. We further validate LiFlow using the refinement network from Nunes
\etal~\cite{nunes2024scaling} to postprocess the generated complete scenes $X_1$ and scale up the number of points by a factor $F=6$, \ie, $X_1^\dagger \in \mathbb{R}^{O \times 3}, O=FM$. We use the provided weights also trained on SemanticKITTI and its default parameters.

\subsubsection{Metrics}
We evaluate and compare LiFlow using the Chamfer Distance (CD), the Jensen-Shannon Divergence (JSD) and the Voxel Intersection over Union (Voxel IoU).
The Chamfer Distance (CD) and Jensen-Shannon Divergence (JSD) are two common metrics to evaluate the reconstruction fidelity of 3D point clouds~\cite{nunes2024scaling, xiong2023learning,ran2024towards}. The CD evaluates the completion at the point level, measuring the level of detail of the reconstructed scene by calculating how far its points are from the ground truth scene. The JSD is a statistical metric that compares the point distribution between the reconstruction and the ground truth scene. For the JSD, we follow the evaluation proposed by Xiong
\etal~\cite{xiong2023learning}, where the scene is first voxelized with a grid resolution of $0.5$~m and projected to a Bird’s Eye View (BEV). The Voxel IoU evaluates the occupancy of the reconstructed scene compared with the ground truth scene after voxelizing them. We follow the evaluation proposed by Song \etal~\cite{song2017semantic} and classify the voxel occupancy at three different voxel resolutions, \ie, $0.5$~m, $0.2$~m, and $0.1$~m. While using a voxel resolution of $0.5$~m, the Voxel IoU evaluates the occupancy over the coarse scene, for decreasing voxel resolutions, more fine-grained details are considered.

\subsection{Scene Completion}

\Cref{tab:01} compares LiFlow with state-of-the-art methods on the SemanticKITTI~\cite{behley2019semantickitti} validation set. Point-level methods, including LiDiff~\cite{nunes2024scaling}, LiDPM~\cite{martyniuk2025lidpm}, and LiFlow, outperform voxel-level approaches (LMSCNet~\cite{roldao2020lightweight}, LODE~\cite{li2023lode}, MID~\cite{vizzo2022make}, PVD~\cite{Zhou_2021_ICCV}) in Chamfer Distance (CD). With the refinement network~\cite{nunes2024scaling}, LiFlow achieves the best CD results and leads in Jensen-Shannon Divergence (JSD), demonstrating superior reconstruction fidelity. Our method also delivers competitive performance in scene occupancy, as measured by Voxel IoU. 

\begin{figure*}[!b]
\centering
\includegraphics[width=.8\linewidth]{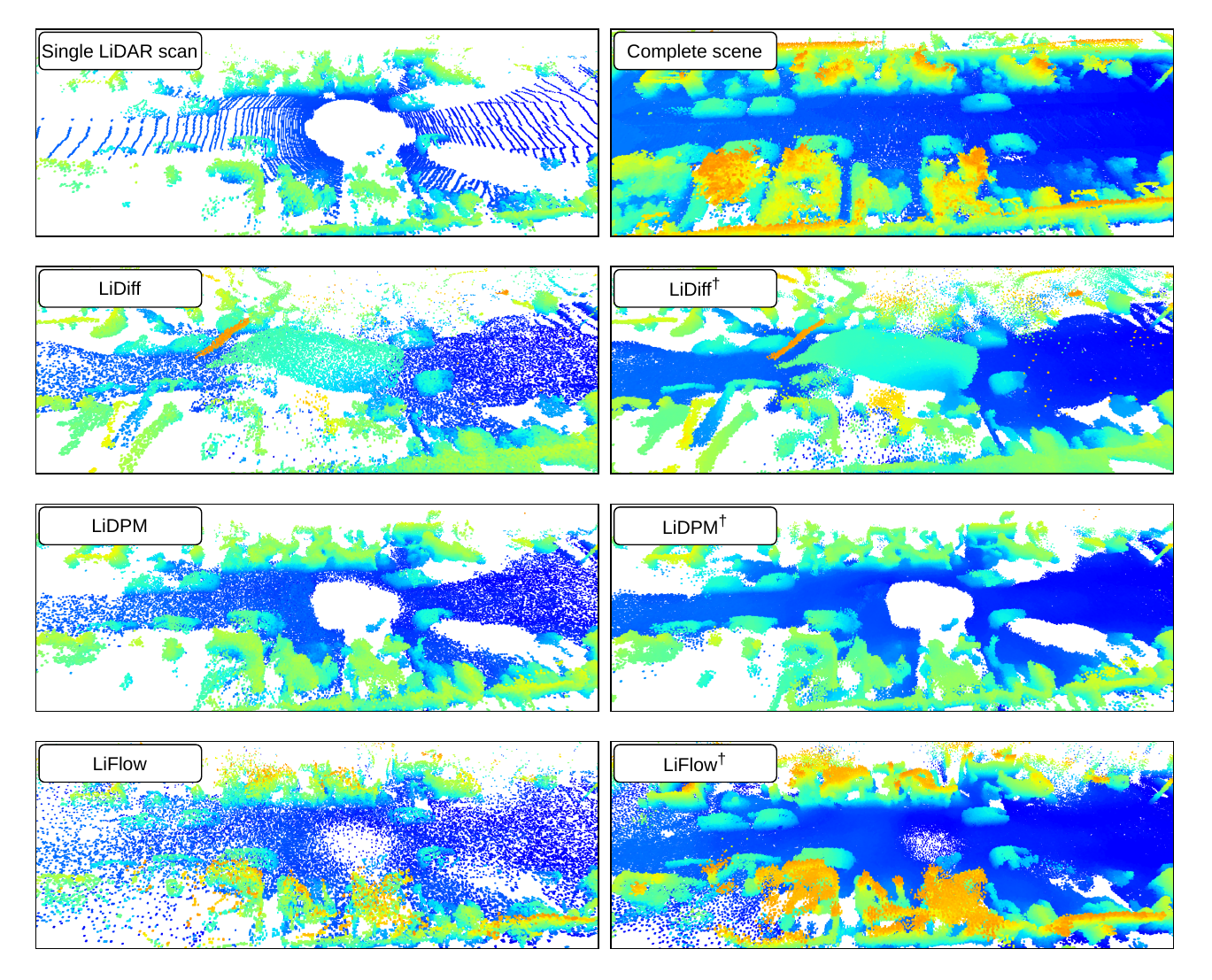}
\caption[Qualitative results on a single LiDAR scan from the SemanticKITTI dataset (sequence $00$).]{Qualitative results on a single LiDAR scan from the SemanticKITTI dataset (sequence $08$). Colors depict point height. $\dagger$: with refinement network~\cite{nunes2024scaling}.}
\label{fig:02}
\end{figure*}

\Cref{fig:02} shows a qualitative comparison of LiFlow with LiDiff~\cite{nunes2024scaling} and LiDPM~\cite{martyniuk2025lidpm} on a single SemanticKITTI~\cite{behley2019semantickitti} LiDAR scan (sequence $08$, $50$ generation steps). With the refinement network~\cite{nunes2024scaling}, LiFlow exhibits a more stable reconstruction and enhanced geometric completion of the scene compared to the other methods, thanks to the consistent distribution between training and inference.

\begin{table}[t]
\caption{Evaluation on Chamfer Distance, Jensen-Shannon Divergence, and Voxel Intersection over Union at different voxel resolutions on the validation set of the SemanticKITTI dataset (sequence $08$). The best and second-best are in bold and underlined, respectively. $\dagger$: with refinement network~\cite{nunes2024scaling}.} 
\centering
\scalebox{1.}{
\begin{tabular}{lccccc}
\hline & \\[-2.ex]
Method & CD[m]$\downarrow$ & JSD[m]$\downarrow$ &\multicolumn{3}{c}{Voxel IoU[\%]$\uparrow$} \\
\cmidrule(lr){4-6}
\\[-2.ex]
&&& $0.5$ & $0.2$ & $0.1$ \\
\hline
\\[-2.ex]
LMSCNet & $0.641$ & $0.431$ & $30.8$ & $12.1$ & $3.7$ \\
LODE & $1.029$ & $0.451$ & $33.8$ & $16.4$ & $5.0$ \\
MID & $0.503$ & $0.470$ & $31.6$ & $22.7$ & $13.1$ \\
PVD & $1.256$ & $0.498$  & $15.9$ & $4.0$ & $0.6$\\
LiDiff & $0.434$ & $0.444$ & $31.5$ & $16.8$ & $4.7$ \\
LiDiff$^\dagger$ & $0.375$ & $0.416$ & $32.4$ & $23.0$ & $\underline{13.4}$ \\
LiDPM & $0.446$ & $0.440$ & $34.1$ & $19.5$ & $6.3$ \\
LiDPM$^\dagger$ & $0.377$ & $\underline{0.403}$ & $\underline{36.6}$ & $\mathbf{25.8}$ & $\mathbf{14.9}$ \\
LiFlow & $\underline{0.309}$ & $0.416$ & $31.6$ & $13.1$ & $3.8$ \\
LiFlow$^\dagger$ & $\mathbf{0.228}$ & $\mathbf{0.367}$ & $\mathbf{37.2}$ & $\underline{25.1}$ & $12.9$ \\
\hline
\\[-2.ex]
 
\end{tabular}
}
\label{tab:01}
\end{table}
\noindent

\Cref{tab:02} compares LiFlow with LiDiff~\cite{nunes2024scaling} and LiDPM~\cite{martyniuk2025lidpm} on Apollo~\cite{lu2019l3} sequence $00$ across different generation steps ($5$, $10$, $20$, $50$) using Euler’s method~\cite{lipman2023flow} and DPM-Solver~\cite{NEURIPS2022_260a14ac}. LiFlow demonstrates superior generalization to unseen environments and stable performance across all step counts. This advantage arises from our flow matching formulation, which aligns the target distribution with the Optimal Transport (OT) path to accelerate convergence. By achieving state-of-the-art performance in fewer steps, LiFlow drastically reduces inference time, facilitating its deployment in autonomous driving applications.

\begin{figure*}[tb]
\centering
\includegraphics[width=.8\linewidth]{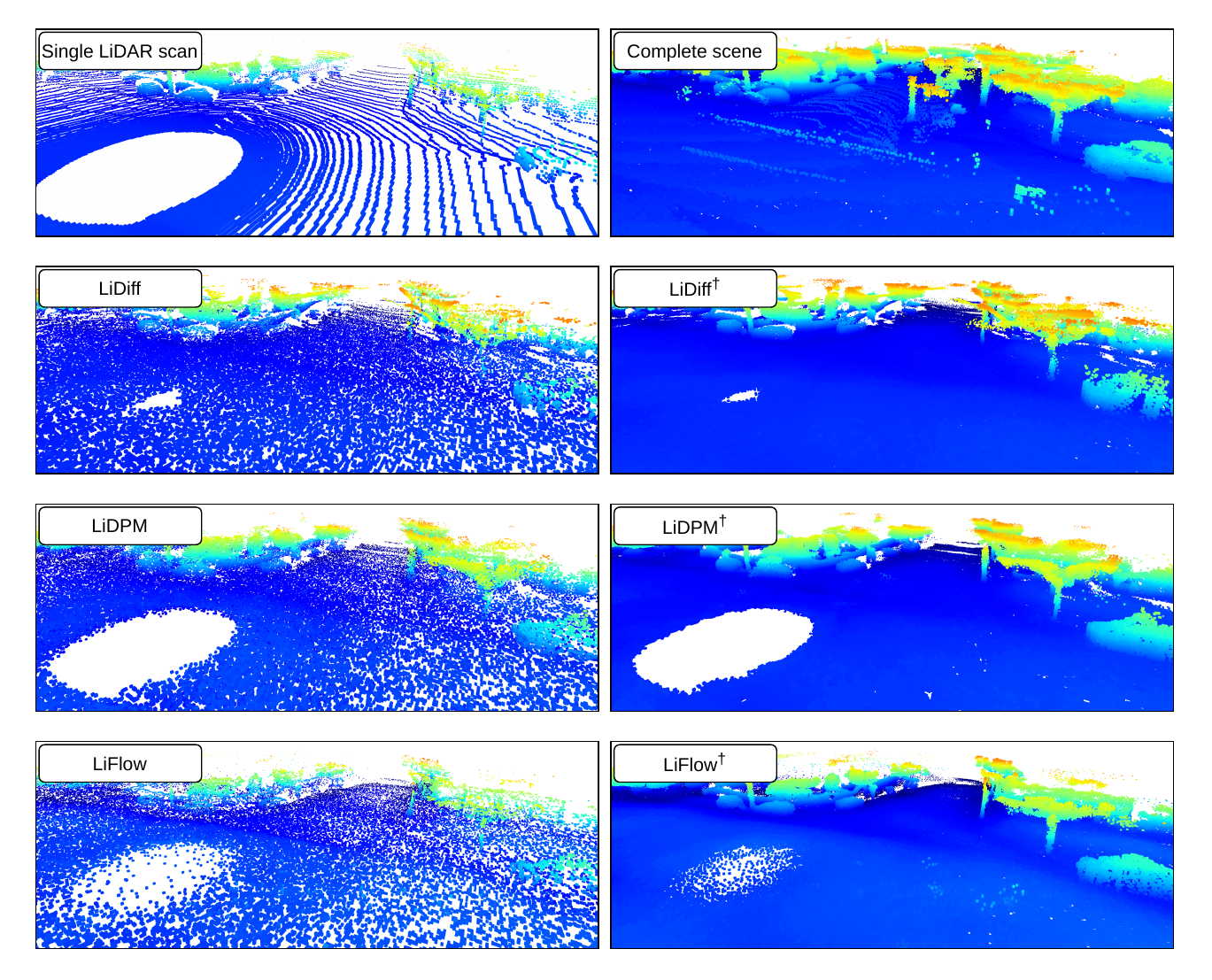}
\caption{Qualitative results on a single LiDAR scan from the Apollo dataset (sequence $00$). Colors depict point height. $\dagger$: with refinement network~\cite{nunes2024scaling}.}
\label{fig:03}
\end{figure*}

\Cref{fig:03} shows a qualitative comparison of LiFlow with LiDiff~\cite{nunes2024scaling} and LiDPM~\cite{martyniuk2025lidpm} on a single Apollo~\cite{lu2019l3} LiDAR scan (sequence $00$, $50$ generation steps). With the refinement network~\cite{nunes2024scaling}, LiFlow achieves superior geometric completion, e.g., of cars, demonstrating state-of-the-art scene reconstruction performance.

\begin{table}[t]
\caption{Evaluation on Chamfer Distance, Jensen-Shannon Divergence, and Voxel Intersection over Union at different voxel resolutions for different generation steps on the Apollo dataset (sequence $00$). Inference time is also reported. The best and second-best are in bold and underlined, respectively. $\dagger$: with refinement network~\cite{nunes2024scaling}.}
\centering
\scalebox{1.}{
\begin{tabular}{lccccccc}
\hline & \\[-2.ex]
Method & Steps & CD[m]$\downarrow$ & JSD[m]$\downarrow$ &\multicolumn{3}{c}{Voxel IoU[\%]$\uparrow$} & Time[s] \\
\cmidrule(lr){5-7}
\\[-2.ex]
&&&& $0.5$ & $0.2$ & $0.1$ \\
\hline
\\[-2.ex]
\multirow{3}{*}{LiDiff}
& $5$ & $0.492$ & $\mathbf{0.381}$ & $21.3$ & $5.4$ & $1.1$ & $5.1$\\
& $10$ & $\mathbf{0.437}$ & $\underline{0.398}$ & $\mathbf{26.7}$ & $9.2$ & $2.3$ & $7.4$\\
& $20$ & $0.457$ & $0.427$ & $\underline{26.3}$ & $\underline{11.3}$ & $\underline{3.3}$ & $12.3$\\
& $50$ & $\underline{0.451}$ & $0.430$ & $25.6$ & $\mathbf{13.4}$ & $\mathbf{4.4}$ & $27.1$\\
\hline
\\[-2.ex]
\multirow{3}{*}{LiDiff$^\dagger$}
& $5$ & $0.396$ & $\mathbf{0.322}$ & $27.6$ & $12.1$ & $3.7$ & $5.2$\\
& $10$ & $\mathbf{0.350}$ & $\underline{0.343}$ & $\mathbf{33.4}$ & $17.8$ & $6.7$ & $7.5$\\
& $20$ & $\underline{0.386}$ & $0.380$ & $\underline{31.3}$ & $\underline{19.0}$ & $\underline{9.2}$ & $12.4$\\
& $50$ & $0.389$ & $0.381$ & $29.5$ & $\mathbf{20.2}$ & $\mathbf{11.8}$ & $27.2$\\
\hline
\\[-2.ex]
\multirow{3}{*}{LiDPM}
& $5$ & $\underline{0.485}$ & $0.442$ & $27.5$ & $14.4$ & $\mathbf{4.7}$ & $5.1$\\
& $10$ & $\mathbf{0.478}$ & $0.440$ & $28.1$ & $14.8$ & $\mathbf{4.7}$ & $7.4$ \\
& $20$ & $\mathbf{0.478}$ & $\underline{0.439}$ & $\underline{28.7}$ & $\underline{15.1}$ & $\underline{4.6}$ & $12.3$\\
& $50$ & $\mathbf{0.478}$ & $\mathbf{0.438}$ & $\mathbf{29.1}$ & $\mathbf{15.3}$ & $\underline{4.6}$ & $27.1$\\
\hline
\\[-2.ex]
\multirow{3}{*}{LiDPM$^\dagger$}
& $5$ & $0.417$ & $\underline{0.405}$ & $30.3$ & $21.0$ & $12.3$ & $5.2$\\
& $10$ & $\mathbf{0.408}$ & $\mathbf{0.403}$ & $\underline{30.8}$ & $\underline{21.4}$ & $\underline{12.5}$ & $7.5$ \\
& $20$ & $\underline{0.409}$ & $\mathbf{0.403}$ & $\mathbf{30.9}$ & $\mathbf{21.6}$ & $\mathbf{12.6}$ & $12.4$\\
& $50$ & $0.410$ & $\mathbf{0.403}$ & $\mathbf{30.9}$ & $\mathbf{21.6}$ & $\mathbf{12.6}$ & $27.2$\\
\hline
\\[-2.ex]
\multirow{3}{*}{LiFlow}
& $5$ & $\mathbf{0.351}$ & $\mathbf{0.398}$ & $\mathbf{28.2}$ & $\mathbf{10.7}$ & $\mathbf{3.0}$ & $4.7$\\
& $10$ & $\underline{0.359}$ & $\underline{0.406}$ & $\underline{27.2}$ & $\underline{10.4}$ & $\underline{2.8}$ & $7.3$\\
& $20$ & $0.365$ & $0.409$ & $26.8$ & $10.0$ & $2.7$ & $12.7$\\
& $50$ & $0.367$ & $0.412$ & $26.6$ & $9.8$ & $2.6$ & $28.8$\\
\hline
\\[-2.ex]
\multirow{3}{*}{LiFlow$^\dagger$}
& $5$ & $\mathbf{0.277}$ & $\mathbf{0.346}$ & $\mathbf{33.4}$ & $\mathbf{21.6}$ & $\underline{11.0}$ & $4.8$\\
& $10$ & $\underline{0.282}$ & $\underline{0.350}$ & $\underline{32.9}$ & $\underline{21.4}$ & $\mathbf{11.1}$ & $7.4$\\
& $20$ & $0.285$ & $0.351$ & $32.6$ & $21.1$ & $10.8$ & $12.8$\\
& $50$ & $0.286$ & $0.352$ & $32.4$ & $20.8$ & $10.6$ & $28.9$\\
\hline
\\[-2.ex]
 
\end{tabular}
}
\label{tab:02}
\end{table}

\subsection{Ablation Studies}

\begin{table}[!t]
\caption{Ablation study on LiFlow for the different loss weights ($\lambda_{\text{NFM}}$, $\lambda_{\text{CDM}}$) on the Apollo dataset (sequence $00$). The best and second-best are in bold and underlined, respectively. $\dagger$: with refinement network~\cite{nunes2024scaling}.}
\centering
\scalebox{1.}{
\begin{tabular}{ccc|ccccc}
\hline & \\[-2.ex]
 \multicolumn{3}{c}{} & CD[m]$\downarrow$ & JSD[m]$\downarrow$ &\multicolumn{3}{c}{Voxel IoU[\%]$\uparrow$} \\
\cmidrule(lr){6-8}
\\[-2.ex]
$\lambda_{\text{NFM}}$ & $\lambda_{\text{CDM}}$ & \multicolumn{1}{c}{} &&& $0.5$ & $0.2$ & $0.1$ \\
\hline
\\[-2.ex]
$1$ & $0$ && $0.411$ & $0.414$ & $29.0$ & $10.5$ & $2.8$ \\
$1$ & $0$ & $\dagger$ & $0.325$ & $\underline{0.356}$ & $\mathbf{34.8}$ & $\mathbf{22.3}$ & $\underline{10.6}$ \\
$0$ & $1$ && $0.386$ & $0.417$ & $21.6$ & $7.7$ & $2.4$ \\
$0$ & $1$ & $\dagger$ & $0.323$ & $0.397$ & $25.3$ & $13.8$ & $6.7$ \\
$1$ & $1$ && $0.355$ & $0.397$ & $26.5$ & $9.7$ & $2.7$ \\
$1$ & $1$ & $\dagger$ & $\underline{0.285}$ & $0.364$ & $31.4$ & $19.7$ & $10.0$ \\
\hline
\\[-2.ex]
$1$ & $0.1$ && $0.359$ & $0.406$ & $27.2$ & $10.4$ & $2.8$ \\
$1$ & $0.1$ & $\dagger$ & $\mathbf{0.282}$ & $\mathbf{0.350}$ & $\underline{32.9}$ & $\underline{21.4}$ & $\mathbf{11.1}$ \\

\hline
\\[-2.ex]

\end{tabular}
}
\label{tab:03}
\end{table}

\Cref{tab:03} presents an ablation study on the loss weights $\lambda_{\text{NFM}}$ and $\lambda_{\text{CDM}}$ on the Apollo\cite{lu2019l3} sequence $00$. Using only Nearest Neighbor Flow Matching (NFM) increases the Chamfer Distance (CD) due to the lack of direct CD supervision, coarser voxel IoU ($0.5$, $0.2$) improves, but fine-grained occupancy ($0.1$) drops as point dispersion is not encouraged. Using only Chamfer Distance Matching (CDM) reduces overall performance, particularly Voxel IoU, since NFM constitutes the foundation of the iterative generative process. Equal weighting of both losses yields performance comparable to the baseline, with CDM encouraging greater point spread, highlighting the complementary roles of NFM and CDM in the proposed FM formulation. 

\Cref{tab:04} presents an ablation study on the MinkUNet~\cite{choy20194d} $u_\theta$ of LiFlow with Batch Normalization (BN)~\cite{ioffe2015batch} layers and Instance Normalization (IN)~\cite{ulyanov2016instance} layers on the Apollo\cite{lu2019l3} sequence $00$. We observe that the use of IN layers significantly improves overall performance across all metrics. In particular, the increased stability during training and inference~\cite{martyniuk2025lidpm} leads to a substantial enhancement in the occupancy of the reconstructed scenes.

\begin{table}[tb]
\caption{Ablation study on the MinkUNet $u_\theta$ of LiFlow with Batch Normalization (BN) layers and Instance Normalization (IN) layers on the Apollo dataset (sequence $00$). $\dagger$: with refinement network~\cite{nunes2024scaling}.}
\centering
\scalebox{1.}{
\begin{tabular}{lccccc}
\hline & \\[-2.ex]
Method & CD[m]$\downarrow$ & JSD[m]$\downarrow$ &\multicolumn{3}{c}{Voxel IoU[\%]$\uparrow$} \\
\cmidrule(lr){4-6}
\\[-2.ex]
&&& $0.5$ & $0.2$ & $0.1$ \\
\hline
\\[-2.ex]
BN & $0.405$ & $\mathbf{0.397}$ & $24.9$ & $7.7$ & $1.8$ \\
IN & $\mathbf{0.359}$ & $0.406$ & $\mathbf{27.2}$ & $\mathbf{10.4}$ & $\mathbf{2.8}$ \\
\hline
\\[-2.ex]
BN$^\dagger$ & $0.340$ & $0.354$ & $29.0$ & $16.3$ & $7.0$ \\
IN$^\dagger$ & $\mathbf{0.282}$ & $\mathbf{0.350}$ & $\mathbf{32.9}$ & $\mathbf{21.4}$ & $\mathbf{11.1}$ \\
\hline
\\[-2.ex]
 
\end{tabular}
}
\label{tab:04}
\end{table}

\section{Conclusion}
We proposed LiFlow as an innovative flow matching formulation that aims to overcome the limitations of the current diffusion methods. We demonstrated its effectiveness through extensive experiments, achieving state-of-the-art results across multiple metrics. While LiFlow achieves consistent improvements in CD, JSD, and surface coverage, its Voxel IoU at fine-grained resolution ($0.1$) is slightly lower than diffusion-based baselines. This behavior reflects a trade-off between global geometric consistency and fine-grained occupancy precision, which is inherent to the proposed nearest neighbor flow matching. Improving fine-resolution occupancy while preserving global consistency is an interesting direction for future work.

%
%
\bibliographystyle{splncs04}
\bibliography{main}
\end{document}